\documentclass[10pt,twocolumn,letterpaper]{article}

\usepackage[pagenumbers]{cvpr} %

\newcommand{\methodname}{Prometheus}
\newcommand{\method}{\texttt{\methodname}\xspace}
\newcommand*{\affaddr}[1]{#1} 
\newcommand*{\affmark}[1][*]{\textsuperscript{#1}}
\renewcommand{\thefootnote}{\fnsymbol{footnote}}

\newcommand\nonumfootnote[1]{%
\begingroup%
    \renewcommand\thefootnote{}\footnote{\hspace{-3.7pt}#1}%
    \addtocounter{footnote}{-1}%
\endgroup%
}

\usepackage{multirow} 
\usepackage{makecell}

\usepackage{colortbl}

\definecolor{firstcolor}{rgb}{1, 0.6, 0.6}
\definecolor{secondcolor}{rgb}{1, 0.8, 0.6}
\definecolor{thirdcolor}{rgb}{1,1, 0.6}

\newcommand{\fst}[1]{\cellcolor{firstcolor}#1}
\newcommand{\snd}[1]{\cellcolor{secondcolor}#1}
\newcommand{\trd}[1]{\cellcolor{thirdcolor}#1}

\definecolor{cvprblue}{rgb}{0.21,0.49,0.74}
\usepackage[pagebackref,breaklinks,colorlinks,allcolors=cvprblue]{hyperref}

\def\paperID{XXXX} %
\def\confName{CVPR}
\def\confYear{2025}

\title{Prometheus: 3D-Aware Latent Diffusion Models for Feed-Forward \\ Text-to-3D Scene Generation}
\author{
Yuanbo Yang\affmark[1*] \quad
Jiahao Shao\affmark[1*] \quad
Xinyang Li\affmark[2] \quad 
Yujun Shen\affmark[3] \quad 
Andreas Geiger\affmark[4] \quad
Yiyi Liao\affmark[1$\dagger$]
\vspace{0.5em}\\
\affaddr{\affmark[1]Zhejiang University}  \quad  \affaddr{\affmark[2]Xiamen University}  \quad  \affaddr{\affmark[3]Ant Group}  \quad  \affaddr{\affmark[4]University of Tübingen} 
}

\newcommand{\by}{\mathbf{y}}

\newcommand{\bI}{\mathbf{I}}

\newcommand{\bp}{\mathbf{p}}
\newcommand{\bt}{\mathbf{t}}

\newcommand{\bc}{\mathbf{c}}
\newcommand{\bm}{\mathbf{m}}

\newcommand{\bd}{\mathbf{d}}
\newcommand{\br}{\mathbf{r}}

\newcommand{\cN}{\mathcal{N}}

\newcommand{\cL}{\mathcal{L}}

\makeatletter
\DeclareRobustCommand\onedot{\futurelet\@let@token\@onedot}
\def\@onedot{\ifx\@let@token.\else.\null\fi\xspace}

\makeatother

\renewcommand{\eqref}[1]{Eq.~\ref{#1}}

\newcommand{\boldparagraph}[1]{\vspace{0.2cm}\noindent{\bf #1:} }

\newif\ifcomment
\commenttrue
\ifcomment
	\newcommand{\yl}[1]{ \noindent {\color{cyan} {\bf Yiyi:} {#1}} }
	
\else
	\newcommand{\ag}[1]{}
	\newcommand{\yl}[1]{}
\fi

\begin{document}
\twocolumn[{%
\renewcommand\twocolumn[1][]{#1}%
\maketitle
\begin{center}
  {
  \vspace{-5pt}
  \captionsetup{type=figure}
  \centering
  \includegraphics[width=0.90\linewidth]{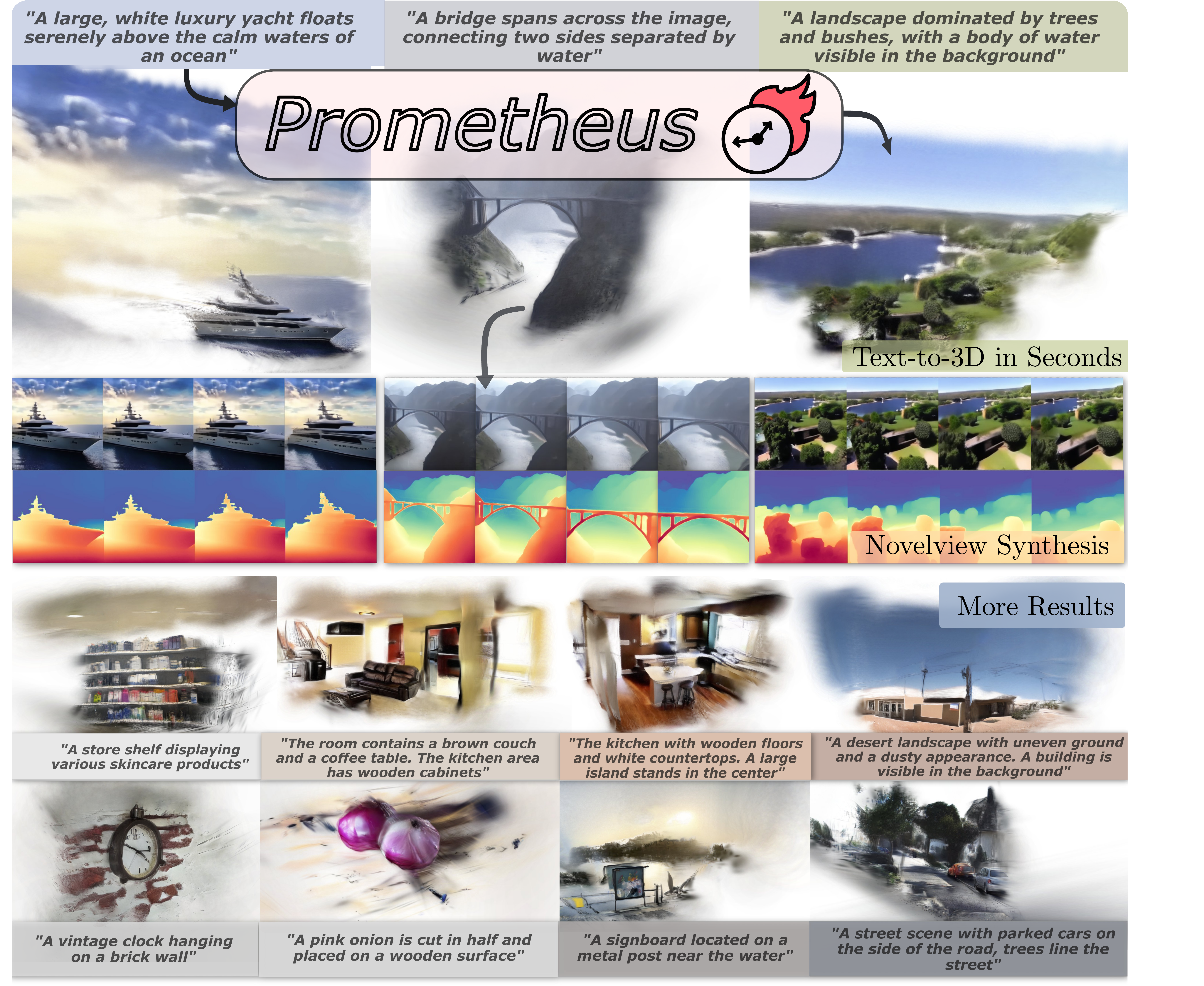}
  \vspace{-8pt}
  \captionof{figure}{
  \textbf{We present \method{}, a novel method for feed-forward scene-level 3D generation.} At its core, our approach harnesses the power of 2D priors to fuel generalizable and efficient 3D synthesis -- hence our name, \method{}.
  }
  \label{fig:teaser}
  \vspace{8pt}
}
\end{center}
}]

\begin{abstract}
    \nonumfootnote{$\ast$ denotes equal contribution.}%
    \nonumfootnote{$\dagger$ corresponding author.}%

In this work, we introduce \method{}, a 3D-aware latent diffusion model for text-to-3D generation at both object and scene levels in seconds. 
We formulate 3D scene generation as multi-view, feed-forward, pixel-aligned 3D Gaussian generation within the latent diffusion paradigm. 
To ensure generalizability, we build our model upon pre-trained text-to-image generation model with only minimal adjustments, and further train it using a large number of images from both single-view and multi-view datasets.
Furthermore, we introduce an RGB-D latent space into 3D Gaussian generation to disentangle appearance and geometry information, enabling efficient feed-forward generation of 3D Gaussians with better fidelity and geometry.
Extensive experimental results demonstrate the effectiveness of our method in both feed-forward 3D Gaussian reconstruction and text-to-3D generation. 
Project page:
\href{https://freemty.github.io/project-prometheus/}{freemty.github.io/project-prometheus}.

\end{abstract}
    
\section{Introduction}
\label{sec:intro}

3D assets play a crucial role in a wide range of applications, including AR/VR, gaming, and simulation. Developing 3D generative models capable of efficiently producing versatile 3D content has become a key objective, drawing substantial interest in the field. Despite rapid progress in 2D image and video generation, 3D generative models continue to fall short of the progress seen in 2D generation.

One line of existing 3D generative models learn from 3D/multi-view data~\cite{xu2023dmv3d,chen2024LaRa}, or single-view images of a single category~\cite{Schwarz2020graf,Chan2021eg3d}. This allows them for directly learning 3D representations in a feed-forward manner. Despite achieving excellent multi-view consistency with good geometry, their \textbf{generalizability} is limited due to the scarcity of the training data.
Another line of approaches seeks to use models trained on large amounts of 2D data for 3D generation –– while the largest multi-view datasets contain around 100K samples, single-view datasets and pre-trained models based on them can reach scales of 100M to 2B samples.
Most methods in this area obtain 3D representations through optimization. Some approaches use score distillation~\cite{poole2022dreamfusion,lin2023magic3d,wang2023prolificdreamer,tang2023dreamgaussian,yi2023gaussiandreamer} or incremental inpainting~\cite{yu2024wonderjourney,wang2024vistadream,shriram2024realmdreamer,chung2023luciddreamer}. However, since 2D models lack a complete understanding of 3D, their outputs sometimes face the Janus problem and tend to produce results with low \textbf{fidelity}.  Another set of methods fine-tunes 2D models to generate multi-view images, requiring multi-view reconstruction to form the 3D representation~\cite{shi2023MVDream,liu2023syncdreamer,gao2024cat3d,sargent2023zeronvs,wu2023reconfusion}. In both cases, the optimization process can be time-consuming, thereby lacking \textbf{efficiency}.

To address the aforementioned issues, 
we introduce \method{}, a 3D-aware latent diffusion model tailored for text-to-3D generation at both object and scene levels. 
Our key idea is to exploiting vast amount of 2D data as well as 2D generative models to facilitate feed-forward 3D generation while maintaining generalization ability -- taming the fire of 2D priors to streamline 3D generation.
Specifically, we formulate 3D scene generation as multi-view, feed-forward, pixel-aligned 3D Gaussian generation within latent diffusion paradigm. To ensure generalizability, we not only build our model upon pre-trained text-to-image generation model (Stable Diffusion) with only minimal adjustments but also train it using both single-view images and multi-view images.
Furthermore, we introduce an RGB-D latent space into 3D Gaussian generation to disentangle appearance and geometry information, enabling efficient feed-forward generation of 3D Gaussians with better fidelity and geometry.

Following the standard latent diffusion paradigm~\cite{rombach2021highresolution}, we separate training into two distinct stages.
In the \textit{first stage}, we train a 3D Gaussian Variational Autoencoder (GS-VAE) that takes multi-view or single-view RGB-D images as input and predicts per-pixel aligned 3D Gaussians. Here, the input depth map during training is estimated using an off-the-shelf monocular depth estimator. Additionally, the encoder of our GS-VAE directly re-uses the Stable Diffusion encoder, predicting latent codes for both RGB images and depth maps. 
We subsequently train a multi-view GS decoder to generate multi-view 3D Gaussians from the RGB-D latent codes conditioned on camera poses.
In the \textit{second stage}, we train a multi-view LDM that jointly predicts multi-view RGB-D latent codes, conditioned on both camera pose and text prompt.
Furthermore, our full model is trained on a combination of 9 multi-view and single-view datasets, aiming for generalizability comparable to Stable Diffusion. 
We demonstrate the effectiveness of our method in both feed-forward 3D Gaussian reconstruction and text-to-3D generation, showcasing that our model can generate 3D scenes in seconds while generalizing well to a variety of 3D objects and scenes.

\section{Related work}
\label{sec:related}

\boldparagraph{3D Generative Models}
3D generative models have attracted great attention in recent years. Many approaches learn from single-category images or 3D supervisions~\cite{Schwarz2020graf, Chan2021eg3d, muller2023diffrf, zhang2024gaussiancube,chen2024microdreamer,wang2024geco}, utilizing either GANs or diffusion models. However, these studies primarily concentrate on domain-specific, object-centric scenes, such as those involving Carla Cars~\cite{Dosovitskiy2017CARLA} and human faces~\cite{Karras2019stylegan}.
Despite recent progress that has further extended these methods to model scene-level generation~\cite{Niemeyer2021CVPR,Xu2023CVPR,Bahmani2023,yang2023iccv}, these methods are yet confined to a specific domain with limited generalization capability. This is mainly due to the scarcity of 3D supervision. We aim to tackle this problem by combining multi-view supervision with a vast amount of 2D images.

\boldparagraph{3D Generation with 2D priors}
Thanks to the rapid progress in 2D generation models like Stable Diffusion~\cite{rombach2021highresolution} and SoRA~\cite{sora}, there is a large many works have explored the potential of large 2D diffusion models for 3D-aware generation. 
One line of works fine-tune 2D diffusion models to enable pose controllability for objects~\cite{liu2023zero1to3,liu2023syncdreamer} or scenes~\cite{sargent2023zeronvs,gao2024cat3d,wang2024motionctrl, he2024cameractrl},
where the output still lies in the 2D image space.
Additionally, some works~\cite{muller2024multidiff, liu2024reconx, yu2024viewcrafter,chen2024v3d,you2024nvssolver,kwak2024vivid123,sun2024dimensionxcreate3d4d,zhao2024genxd,kwak2024vivid123} aim to incorporate underlying 3D prior knowledge to assist in synthesis. 
Since the aforementioned methods generate only 2D images, a separate 3D reconstruction step is still needed, which can add time and introduce errors. In contrast, we explore directly generating 3D representations in a feed-forward manner.

Another line of methods utilize the 2D priors for 3D generation through optimization, e.g., using Score Distillation Sampling (SDS)~\cite{poole2022dreamfusion,poole2022dreamfusion, wang2022sjc, wang2023prolificdreamer, tang2023dreamgaussian, yi2023gaussiandreamer}.
In parallel, several works~\cite{fridman2023scenescape, lei2023rgbd2, yu2024wonderworld, zhang20243DitScene, chung2023luciddreamer,wang2024vistadream,shriram2024realmdreamer} formulate scene synthesis as ``perpetual view generation''~\cite{liu2020infinitenature}, synthesizing the effects of navigating a 3D world by stitching and rendering images based on camera motions. 
These methods do not require retraining 2D generative models, but they are inefficient due to per-scene optimization. The generated content's quality is limited by the 2D backbone, causing issues like multi-view inconsistency and artifacts in geometry and texture during image inpainting and score distillation.

\boldparagraph{Feed-Forward 3D Gaussian Generation}
Unlike view synthesis-based methods that focus on synthesizing multi-view images followed by 3D reconstruction, a more intuitive approach is to directly generate 3D representations.
Following this idea, many works~\cite{wang2021ibrnet, chan2023genvs, hong2023lrm, xu2023dmv3d} concentrate on the direct synthesis of 3D representations, such as NeRF~\cite{mildenhall2020nerf}, which generates meshes from single or few-view image inputs.
Recently, numerous works~\cite{charatan23pixelsplat, chen2024mvsplat, wewer24latentsplat, szymanowicz2024flash3d, gslrm2024, xie2024lrmzero, xu2024grm, tang2024lgm, chen2024LaRa, chen2024mvsplat360, xu2024depthsplat,xu2024depthsplat} have adopted 3D Gaussian Splatting~\cite{kerbl3Dgaussians} as the underlying representation. 
PixelSplat~\cite{charatan23pixelsplat} is the first feed-forward model that learns to reconstruct 3D Gaussian splats from pairs of images. GS-LRM~\cite{gslrm2024} builds on this idea and utilizing a larger reconstruction model (LRM~\cite{hong2023lrm}), achieving improved results.

To address this challenge, several works~\cite{Schwarz2024wildfusion, li2024dual3d, lan2024ln3diff, li2024director3d,paul2024samplinginseconds,he2024gvgen,zhou2024DiffGS,roessle2024l3dg} propose Large Diffusion Models as powerful generators for 3D representations. WildFusion~\cite{Schwarz2024wildfusion} introduces a novel approach for achieving 3D-aware image synthesis from in-the-wild datasets using latent diffusion models.
Director3D~\cite{li2024director3d}, which is closely related to our work, presents a robust open-world text-to-3D generation framework designed to create both real-world 3D scenes and adaptive camera trajectories. Unlike Director3D, which requires supervision in the image space, we follow common practices in 2D image generation and adopt a latent diffusion framework. This approach significantly reduces computational overhead, making larger-scale training feasible, and better leverages the 2D latent space, enhancing the generalizability of our method.

\section{Method}
\label{sec:method}
\newcommand{\mat}[1]{\mathbf{#1}}
\newcommand{\set}[1]{\mathcal{#1}}
\newcommand{\real}{\mathbb{R}}
\newcommand{\render}{\texttt{VR}}
\newcommand{\diffrender}{\texttt{R}}
\newcommand{\merge}{\texttt{M}}
\newcommand{\encoder}{\mathcal{E}_\mathbf{\phi}}
\newcommand{\crossviewvit}{\mathcal{C}_\mathbf{\phi}}
\newcommand{\decoder}{\mathcal{D}_\mathbf{\phi}}

\newcommand{\image}{I}
\newcommand{\rgb}{\set{I}}
\newcommand{\depth}{\set{D}}
\newcommand{\pose}{\set{R}}
\newcommand{\raymap}{\set{R}}

\newcommand{\latent}{\mathbf{z}}
\newcommand{\latentrgb}{\latent^{(I)}}
\newcommand{\latentdepth}{\latent^{D}}

\newcommand{\perpixelgs}{F}
\newcommand{\scenegs}{G}

\newcommand{\mvldm}{\mathcal{G}_\mathbf{\theta}}
\newcommand{\unet}{F_\mathbf{\theta}}

\newcommand{\pseudolatent}{\hat{\latent}}
\newcommand{\pseudolatentdepth}{\pseudolatent^{(\depth)}}
\newcommand{\noise}{\bm{\epsilon}}

\newcommand{\denoiser}{\mathbf{s}_\mathbf{\theta}}
\newcommand{\seconddenoiser}{D_\mathbf{\theta}}
\newcommand{\thirddenoiser}{F_\mathbf{\theta}}
\newcommand{\denoiserlong}{\seconddenoiser(\latentdepth_t, \latentrgb, t)}

\begin{figure*}[t]
  \centering
   \includegraphics[width=1.\linewidth]{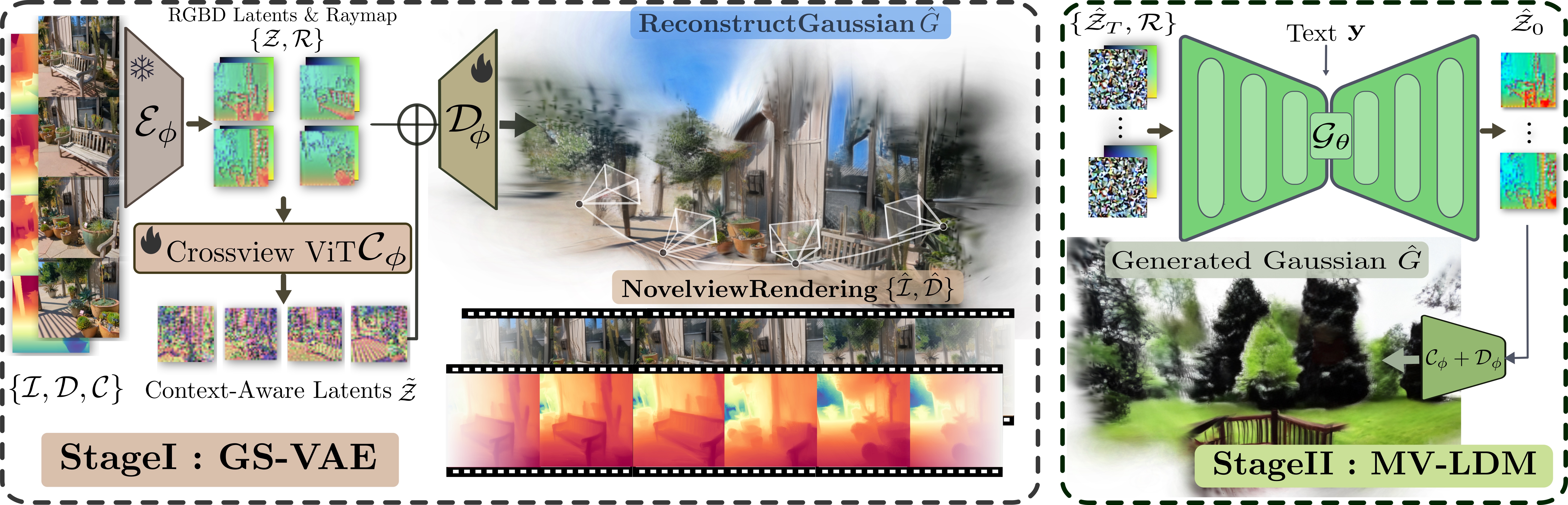}

   \caption{\textbf{Method Overview.} Our training process is divided into two stages. In stage 1, our objective is to train a GS-VAE. Utilizing multi-view images along with their corresponding pseudo depth maps and camera poses, our GS-VAE is designed to encode these multi-view RGB-D images, integrate cross-view information, and ultimately decode them into pixel-aligned 3DGS. In stage 2, we focus on training a MV-LDM. We can generate multi-view RGB-D latents by sampling from randomly-sampled noise with trained MV-LDM.
   }
   \label{fig:method}
\end{figure*}

In this section, we provide the technical details of our method. As illustrated in ~\cref{fig:method}, \method{} follows the common latent diffusion framework~\cite{rombach2021highresolution}, which involves two training stages. In the first stage (\cref{sbusce:gsvae}), our 3D autoencoder, GS-VAE, learns a compressed and abstracted latent space from multi-view images. Subsequently, it decodes this latent space into pixel-aligned 3DGS representations, serving as scene-level representations. In the second stage (\cref{sbusec:mvldm}), a latent multi-view diffusion model (MV-LDM) is trained on the latent representations derived from the first stage's autoencoder. This process results in a fully generative model. Finally, we elaborate on our sampling strategy (\cref{subsec:inference}) for sampling 3D scenes in seconds while maintaining consistency and visual fidelity.

\subsection{Stage 1: GS-VAE}
\label{sbusce:gsvae}

In Stage 1, our objective is to train a 3D autoencoder capable of compressing data into a latent space and subsequently reconstructing it into a 3D representation. Given multi-view input images with camera poses, our GS-VAE outputs multi-view pixel-aligned 3DGS. These outputs are then merged into a scene-level 3D representation.

\noindent\textbf{Encoding Multi-View RGB-D Images.} 
We propose to encode both, RGB images and their predicted monocular depth maps, into the latent space, considering that monocular depth maps provide clues for the later 3D Gaussian decoding process and can be easily obtained. 
Given a set of multi-view images $\set{I} = \left \{ \image_i \in \real^{H \times W \times 3} | i = 1,2,\dots, N \right \}$ with each image $\image_i$ being an observation of an underlying 3D scene, we first employ an off-the-shelf depth estimator~\cite{depth_anything_v2} to obtain their corresponding monocular depth maps $\set{D} = \left \{ D_i \in \real^{H \times W \times 1} | i = 1,2,\dots, N \right \}$.
Next, we utilize a pre-trained image encoder $\encoder$ to encode both the multi-view images $\set{I}$ and their depth maps $\set{D}$ into a latent representation:
\begin{equation}
\encoder: (\set{I}, \set{D}) \mapsto \set{Z} \in \real^{N \times h \times w \times c},
\label{eq:encoder}
\end{equation}
where $h\times w$ is the downsampled resolution.
In practice, we use a pre-trained Stable Diffusion (SD) image encoder and freeze it during training. Recent methods, such as Marigold~\cite{ke2023repurposing}, indicate that the SD encoder exhibits robust generalization capabilities with depth maps. Consequently, we opt to employ the same SD encoder to independently encode both images and depths without the need for fine-tuning. Subsequently, we concatenate these encoded representations to obtain the full multi-view latent $\mathcal{Z}$ which can be used for 3D reconstruction. Our diffusion model is additionally trained within this joint RGB-D latent space.

\noindent\textbf{Fusing Multi-View Latent Images.}
Recent advancements~\cite{zhang2024raydiffusion, gslrm2024, lan2024ln3diff, Wang2024dust3r, hong2023lrm} underscore the significant potential of transformer-based models in integrating multi-view information. Since our latent codes for each view in $\set{Z}$ are derived individually, we employ a multi-view transformer to facilitate cross-view information exchange.

We further inject $N$ camera poses into our multi-view transformer.
Inspired by recent works~\cite{gao2024cat3d,xu2023dmv3d,li2024director3d,lan2024ln3diff}, we choose Plücker coordinates as camera representation~\cite{sitzmann2021lfns}, specifically $\br = (\bd, \bp \times \bd) \in \real^{6}$, where $\bd$ denotes the normalized ray direction and $\bp$ denotes the camera origin. Thus, initial $N$ camera poses can be re-parameterized as multi-view ray maps $\set{R} = \left \{ R_i \in \real^{H \times W \times 6} | i = 1,2,\dots, N \right \}$.
We combine the multi-view latent codes $\set{Z}$ and the camera ray maps $\set{R}$ via concatenation along the feature channel and feed them into the cross-view transformer to obtain the fused latent codes $\tilde{\set{{Z}}}\in \real^{h \times w \times c}$ that merges multi-view context:
\begin{equation}
\crossviewvit: (\set{Z}, \set{R}) \mapsto \tilde{\set{{Z}}} \in \real^{h \times w \times c}.
\label{eq:cross_view_vit}
\end{equation}

\noindent\textbf{Decoding into Gaussian Scenes.}
Finally, we concatenate the raw image latent codes $\set{{Z}}$, ray maps $\set{R}$, and the fused latent codes $\tilde{\set{{Z}}}$, and feed them into the decoder, thereby obtaining the pixel-aligned multi-view 3D Gaussians 
$\set{F} = \left \{ F_i \in \real^{H \times W \times C_G} | i = 1,2,\dots, N \right \}$
\begin{equation}
\label{eq:gs_decoder}
\decoder: (\set{Z}, \tilde{\set{Z}}, \set{R}) \mapsto \set{F} \in \real^{N \times H \times W \times C_G},
\end{equation}
where $F_i$ is the pixel-aligned 3D Gaussians corresponding to each image. 
A 3D Gaussian is parameterized by 1-channel depth, 4-channel rotation quaternion, 3-channel scale, 1-channel opacity, and 3-channel spherical harmonics coefficients respectively.
Thus $C_G=12$ in our formulation.
After aggregating multi-view 3D Gaussians, we can get the final scene-level 3D Gaussians $G$ as in \cref{eq:merge}
\begin{equation}
\label{eq:merge}
\merge(\set{F}) \mapsto G \in \real^{N_G \times C_G}.
\end{equation}
Here, $\merge(\cdot)$ denotes the aggregation operation, which is achieved by transforming all 3D Gaussians into a global coordinate system.
$N_G$ represents the number of full Gaussian primitives, which is equivalent to $N \times H \times W$.

In practice, this architecture is also applicable to single-view images, where $N$ equals to $1$. During training, we sample from both single-view and multi-view images.
Besides, to maximize the usage of 2D generative priors, we follow Director3D~\cite{li2024director3d} and repurpose a pre-trained Stable Diffusion image decoder with minor modifications as our Gaussian decoder $\decoder$. Specifically, we only adjust the number of channels in the first and last convolutional layers.
 
\noindent\textbf{Loss Function.} Given the reconstructed scene-level 3D Gaussians $\hat{\scenegs}$, we can render them from arbitrary viewpoints. Let $\bc$ denote a given viewpoint, we can render the corresponding RGB image and depth map from $\hat{\scenegs}$:
\begin{equation}
\diffrender(\hat{G}, \bc) \mapsto \{ \hat{I}, \hat{D} \},
\label{eq:render}
\end{equation}
where $\diffrender(\cdot)$ denotes the differentiable rendering of 3D Gaussian Splatting. We can subsequently apply render loss, which integrates MSE (Mean Squared Error) loss and perceptual~\cite{johnson2016Perceptualloss} loss:
\begin{equation}
\label{eq:rendering_loss}
\cL_{render} = \cL_{mse}(\hat{I}, I) +  \cL_{vgg}(\hat{I}, I).
\end{equation}

In addition to the rendering loss on the RGB domain, we also impose a loss between our rendered expected depth $\hat{D}$ and the monocular depth $\bar{D}$ which serve as pseudo geometry ground truth as below:
\begin{equation}
\label{eq:depth_loss}
\cL_{depth} =  \lVert (w \hat{D}  + q) - \bar{D}\rVert _2,
\end{equation}
where $\cL_{depth}$ is the scale invariant depth loss following~\cite{Ranftl2022midas}. Here, $w$ and $q$ are the scale and shift used to align $\hat{D}
$ with $\bar{D}$ since $\bar{D}$ is defined only up to a scale and a shift. We determine $w$ and $q$ using a least-squares criterion~\cite{Ranftl2022midas}.  

Our full loss function of the GS-VAE is as follows:
\begin{equation}
\label{eq:gsvae_loss}
\cL(\phi) = \lambda_{1} \cL_{mse} + \lambda_{2} \cL_{vgg} +\lambda_{3} \cL_{depth},
\end{equation}
where $\phi$ denotes the optimizable parameters in GS-VAE, and $\lambda_1, \lambda_2, \lambda_3$ are employed to balance the weight of each loss term.

\subsection{Stage 2: Geometry-Aware Multi-View Denoiser}
\label{sbusec:mvldm}
After the training of stage 1, we obtain a GS-VAE capable of i) efficiently compressing images into a compact latent space and ii) bridging the gap between 2D and 3D via a 3D Gaussian decoder. This capability enables us to train the multi-view diffusion model (MV-LDM) in the latent space.

We aim to jointly generate multi-view RGB-D latent codes $\set{Z} \in \real^{N \times h \times w \times c}$ to provide richer geometric cues for decoding 3DGS. Therefore, we formulate a continuous-time denoising diffusion~\cite{song2020score,karras2022elucidating} conditioned on the text prompt $\by$ and camera poses $\set{R}$. The diffusion model consists of a stochastic \textit{forward} pass to inject one noise level noise Gaussian noise into input latent codes and a \textit{reverse} process to remove noise with a learnable denoiser $\mvldm$.

\noindent\textbf{Training.} 
For each training step, we sample one noise level $\sigma_t$, where $\log\sigma_t \sim \cN(P_{mean}, P_{std}^2)$~\citep{karras2022elucidating}. Next, we add noise of this level to the clean multi-view latents $\set{Z}_0$ to obtain the noisy latents $\set{Z}_t$ as
\begin{equation} 
    \set{Z}_t=\set{Z}_0+\sigma_t^2\boldsymbol{\epsilon}, \ \boldsymbol{\epsilon} \sim \cN(\mathbf{0}, \bI).
\end{equation}
In the reverse process, diffusion model denoises $\set{Z}_t$ towards predicted clean $\hat{\set{Z}}_0$ with a learnable \textit{multi-view denoiser} $\mvldm$ as 
\begin{equation} \label{eq:naive_denoise}
    \hat{\set{Z}}_0=\mvldm(\set{Z}_t; \sigma_t, \by, \set{R}),
\end{equation}
where $\by$ and $\set{R}$ are the text and camera poses condition respectfully.
Our MV-LDM is trained in latent space via~\textit{denoising score matching}~(DSM)~\citep{vincent2011connection}
\begin{equation} \label{eq:naive_diffusion_objective}
    \cL(\theta)=\mathbb{E}_{\set{Z}, \set{R}, \by, \sigma_t} \left[\lambda(\sigma_t) \| \hat{\set{Z}}_0 - \set{Z}_0 \|_2^2 \right],
\end{equation}
with weighting function $\lambda(\sigma)=(1+\sigma^2)\sigma^{-2}$.
In this work, we follow EDM~\cite{karras2022elucidating} and parameterize the denoiser $\mvldm$ as 
\begin{equation} \label{eq:our_edm_preconditioning}
\begin{split}
    \mvldm(\set{Z}_t; \sigma_t, \by, \set{R})& = 
    c_\text{skip}(\sigma_t)\set{Z}_t + \\ c_\text{out}&(\sigma_t) \unet(c_\text{in}(\sigma_t)\set{Z}_t; c_\text{noise}(\sigma_t), \by, \set{R}),
\end{split}
\end{equation}
where $\unet$ is a UNet to be trained in our case, and $c_\text{skip}$, $c_\text{out}$, $c_\text{in}$, and $c_\text{noise}$ are preconditioning functions. Furthermore, consistent with Stage 1~\cref{eq:gs_decoder}, we employ ray maps as the pose representation and incorporate them into the network by concatenating them with the noisy latents $\set{Z}_t$ along feature channel. Additionally, the text prompt conditioning is introduced via cross-attention mechanisms.

Inspired by recent multi-view diffusions~\cite{shi2023MVDream, gao2024cat3d, li2024director3d}, we replace the self-attention blocks in the original UNet with 3D cross-view self-attention blocks to capture multi-view correlations. In practice, to leverage the pre-trained text-to-image prior, we initialize the model $\mvldm$ from a pre-trained text-to-image diffusion model's, specifically the UNet from Stable Diffusion~\cite{rombach2021highresolution}.

\noindent\textbf{Sampling.} 
At sampling time, multi-view latents ${\set{Z}}_0$ is restored from a randomly-sampled Gaussian noise ${\set{Z}}_T$ conditioning on text prompt and camera poses by iteratively applying the denoising process with trained MV-LDM $\mvldm$ 
\begin{equation} \label{eq:initialization}
    {\set{Z}}_T \sim \cN(\mathbf{0}, \sigma_T^2\bI)
\end{equation}
\begin{equation} \label{eq:inference}
\begin{split}
    {\set{Z}}_{t-1} = \frac{{\set{Z}}_{t} - \mvldm({\set{Z}}_T; \sigma_t, \by, \set{R})}{\sigma_{t}} (\sigma_{t-1}-\sigma_{t}) \\+ {\set{Z}}_{t},  \quad 0 < t \le T
\end{split}
\end{equation}
where $\sigma_0,...,\sigma_T$ are sampled from a fixed variance schedule of a denoising process with $T$ steps.

\noindent\textbf{The Importance of Noise Level.} 
Inspired by insights from recent works~\cite{Blattmann_stablevideodiffusion, chen2024v3d, shi2023zero123plus}, we recognize that a lower Signal-to-Noise Ratio (SNR) during the denoising step is crucial for determining the global low-frequency structure of the content. Furthermore, this lower SNR during sampling is essential for achieving multi-view consistency in the multi-view diffusion model $\mvldm$.
Therefore, we adopt a relatively large noise distribution with $P_{mean}$ = 1.5 and $P_{std} = 2.0$ during multi-view training of MV-LDM $\mvldm$ and $P_{mean}$ = -0.5 and $P_{std} = 1.2$ during single-view training.

\begin{table}[t!]
    \centering
    \renewcommand{\tabcolsep}{6pt}
    \resizebox{0.95\linewidth}{!}{
    \begin{tabular}{@{} l|ccc}
    \toprule
    Dataset & Scene type & $\#$ of frames & $\#$ of Scenes \\ 
    \midrule
    SAM-1B~\cite{kirillov2023segany} & Single view & 11M &    
 - \\ 
    MVImgNet~\cite{yu2023mvimgnet} & Object & 6.8M & 230K    \\ 
    DL3DV-10K~\cite{ling2024dl3dv} & Indoor / Outdoor & 2.2M &  6K   \\ 
    Objaverse~\cite{objaverseXL} & Object & 11.5M &  784K   \\ 
    ACID~\cite{liu2020infinitenature} & Indoor & 510K &  11K   \\ 
    RealEstate10K~\cite{zhou2018re10k} & Indoor & 2.8M & 57K    \\ 
    KITTI~\cite{geiger2012we} & Driving & 42K &   0.8K  \\ 
    KITTI-360~\cite{Liao2022PAMI} & Driving & 69K &  1.2K   \\ 
    nuScenes~\cite{caesar2020nuscenes} & Driving & 340K &  0.85K   \\ 
    Waymo~\cite{Sun2020CVPR} & Driving & 200K &  1K   \\ 
    \bottomrule
    \end{tabular}}
    \vspace{-0.3cm}
    \caption{\textbf{Training datasets} We collect a large, multi-domain dataset for training, including single-view and multi-view data, all paired with detailed captions.}
    \vspace{-0.5cm}
\label{tab:datasets}
\end{table}

\subsection{Text to 3D Scene Generation in Sceonds}
\label{subsec:inference}
Based on the model above, we can achieve feed-forward text to 3D scene generation by sampling multi-view RGB-D latents $\set{Z}$ from randomly-sampled Gaussian noise ${\set{Z}}_T$ in latent space using MV-LDM $\mvldm$, and subsequently decode into a 3D Gaussian Scene $G$ using GS-VAE decoder:
\begin{align}
\label{eq:full_generation}
\begin{cases}
\mvldm: ( {\set{Z}}_T ; \by , \set{R}) &\mapsto \set{Z}, \\
\crossviewvit: (\set{Z} , \set{R}) &\mapsto \tilde{\set{Z}}, \\
\decoder: (\set{Z} , \tilde{\set{Z}} , \set{R}) &\mapsto G.
\end{cases}
\end{align}

\begin{table*}[t!]
    \centering
    \renewcommand{\tabcolsep}{6pt}
    \resizebox{0.95\linewidth}{!}{
    \begin{tabular}{@{}l|ccccc|ccccc|ccccc}
    
    \toprule
    \multirow{2}{*}{Method} &  
    \multicolumn{5}{c|}{Tartanair (\textit{Easy})} & 
    \multicolumn{5}{c|}{Tartanair (\textit{Medium})} & 
    \multicolumn{5}{c}{Tartanair (\textit{Hard})} \\
    
    & 
    PSNR$\uparrow$ & SSIM$\uparrow$ & LPIPS$\downarrow$ & 
    \textit{AbsRel}$\downarrow$ & $\delta$1$\uparrow$ & 
    PSNR$\uparrow$ & SSIM$\uparrow$ & LPIPS$\downarrow$ & 
    \textit{AbsRel}$\downarrow$ & $\delta$1$\uparrow$ & 
    PSNR$\uparrow$ & SSIM$\uparrow$ & LPIPS$\downarrow$ & 
    \textit{AbsRel}$\downarrow$ & $\delta$1$\uparrow$  \\
    \midrule
    
    pixelSplat\cite{charatan23pixelsplat} &
    \fst{\textbf{21.65}} & \fst{\textbf{0.681}} & \snd{0.293} & \snd{0.650} & \snd{0.373} &
    \fst{\textbf{20.30}} & \fst{\textbf{0.628}} & \snd{0.337} & \snd{0.806} & \snd{0.323} &
    \snd{19.35} & \fst{\textbf{0.589}} & \snd{0.371} & \snd{0.871} & \snd{0.307} \\
    MVSplat~\cite{chen2024mvsplat} &
    \trd{19.38} & \trd{0.569} & \trd{0.370} & \trd{0.809} & \trd{0.283} &
    \trd{18.50} & \trd{0.531} & \trd{0.409} & \trd{0.872} & \trd{0.265} &
    \trd{17.87} & \trd{0.500} & \trd{0.445} & \trd{0.907} & \trd{0.272} \\                   
    Ours &
    \snd{20.95} & \snd{0.589} & \fst{\textbf{0.289}} & \fst{\textbf{0.435}} & \fst{\textbf{0.536}} &
    \snd{20.15} & \snd{0.560} & \fst{\textbf{0.314}} & \fst{\textbf{0.493}} & \fst{\textbf{0.514}} &
    \fst{\textbf{19.49}} & \snd{0.532} & \fst{\textbf{0.341}} & \fst{\textbf{0.526}} & \fst{\textbf{0.505}} \\
     
    \bottomrule

    \end{tabular}}
    \vspace{-0.3cm}
    \caption{\textbf{Quantitative comparison for Stage 1}. We compare our GS-VAE with baselines for generalizable reconstruction on \textit{Tartanair}.}
    \vspace{-0.3cm}
\label{tab:stage1_comp}
\end{table*}

\begin{figure*}[t!]
  \centering
  \includegraphics[width=0.95\linewidth]{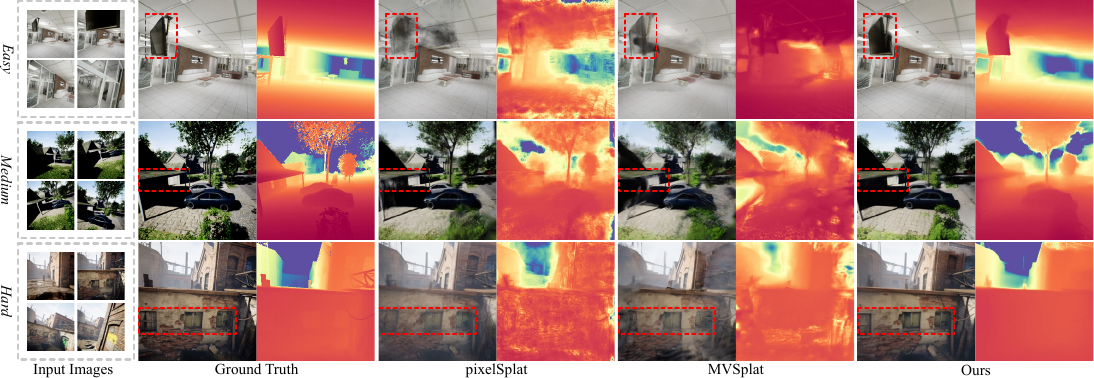}\vspace{-0.1cm}
  \caption{\textbf{Qualitative comparison for Stage 1.} We compare \method{} against baselines under varying difficulty settings. As overlap gradually decreases, the advantages of our method continue to grow. Moreover, as shown in the depth map, our method exhibits superior geometry quality across all settings.}
    \label{fig:stage1_qualitative}
\vspace{-0.3cm}
\end{figure*}

To sample that with high quality and align with the condition, we use classfier-free-guidance (CFG)~\cite{ho2022cfg} to guide multi-view generation toward condition signal $\by$.
\begin{equation} 
\label{eq:cfg}
\begin{split}
\mvldm^{w}(\set{Z}_t;\by, \set{R}) =  w \cdot \mvldm&(\set{Z}_t ; \by , \set{R}) \\ &+ (w - 1) \cdot \mvldm(\set{Z}_t; \set{R}),
\end{split}
\end{equation} 
where $w \ge 0 $ represents the guidance strength. However, if we simply apply the naive CFG~\cref{eq:cfg}, as commonly practiced in most text-to-image methods, increasing $w$ would lead to multi-view inconsistency in the generated results. This naive design described above causes the model to overfit to the text condition while compromising multi-view consistency, in line with with the findings in~\cite{woo2023harmonyview, wu2024direct3d}. To balance multi-view consistency and fidelity during sampling, we follow HarmonyView~\cite{woo2023harmonyview} and adapt hybrid sampling guidance, which rewrites~\cref{eq:cfg} as below:
\begin{align}
\label{eq:harmony_cfg}
\mvldm^{w}(\set{Z}_t; \by, \set{R} ) & =  \mvldm(\set{Z}_t;\by,  \set{R}) \notag \\
+ & w_1 \cdot ( \mvldm(\set{Z}_t ; \by,  \set{R} )  - \mvldm(\set{Z}_t ;  \set{R})) \notag \\
+ & w_2  \cdot ( \mvldm(\set{Z}_t ; \by,  \set{R}  )  - \mvldm(\set{Z}_t ; \by)),
\end{align}
where $w_1$ and $w_2$ denote the weight of text and pose guidance respectively with $w = w_1 + w_2$, thereby better maintaining fidelity and consistency across generated views. In addition, we also use CFG-rescale as proposed in~\cite{lin2024commonareflaw} to avoid over-saturation issues during conditional sampling.

\section{Experiment}
\label{sec:experimenmt}

\subsection{Training Datasets}
\label{subsec:training_dataset}

We train our method on large-scale single-view and multi-view datasets, see~\cref{tab:datasets}. Regarding the single view dataset, we use a high-quality SAM-1B~\cite{kirillov2023segany} dataset with detailed captions~\cite{liu2023llava} which was present in PixArt-$\alpha$~\cite{chen2023pixartalpha}. Our model is trained on a combination of 9 multi-view datasets, including object-centric, indoor, outdoor, and driving scenarios, text prompts for each scene are generated using the multi-modal large language model~\cite{liu2023llava}. 

\begin{table*}[t!]
    \centering
    \renewcommand{\tabcolsep}{6pt}
    \resizebox{0.95\linewidth}{!}{
    \begin{tabular}{@{}l|ccc|ccc|ccc|c@{}}
    
    \toprule
    \multirow{2}{*}{Method} &  
    \multicolumn{3}{c|}{\textit{Single-Object}} & 
    \multicolumn{3}{c|}{\textit{Single-Object-with-Surroundings}} &
    \multicolumn{3}{c|}{\textit{Scene-Level}}  & \multirow{2}{*}{Time} \\
    
    & 
    BRISQUE$\downarrow$ & NIQE$\downarrow$ & CLIP-Score$\uparrow$ & 
    BRISQUE$\downarrow$ & NIQE$\downarrow$ & CLIP-Score$\uparrow$ & 
    BRISQUE$\downarrow$ & NIQE$\downarrow$ & CLIP-Score$\uparrow$ &\\
    \midrule 
    
    GaussianDreamer~\cite{yi2023gaussiandreamer} &
    107.8 & 18.79 & \snd{0.386} &
    110.8 & 18.16 & \snd{0.389} &
    \multicolumn{3}{c|}{-} & $\approx$ 15min\\

    MVDream~\cite{shi2023MVDream}+LGM~\cite{tang2024lgm} &
    \trd{74.64} & \trd{14.96} & \trd{0.379} &
    \trd{77.50} & \trd{14.03} & 0.343 &
    \multicolumn{3}{c|}{-} & \snd{$\approx$ 10s}\\

    Director3D~\cite{li2024director3d} &
    \fst{\textbf{49.91}} & \fst{\textbf{13.56}} & \fst{\textbf{0.397}} &
    \fst{\textbf{49.77}} & \fst{\textbf{13.64}} & \fst{\textbf{0.405}} &
    \snd{50.88} & \snd{14.97} & \snd{0.357} & \trd{$\approx$ 22s} \\

    Ours &
    \snd{59.43} & \snd{14.23} & 0.329 &
    \snd{58.88} & \snd{14.00} & \trd{0.369}  &
    \fst{\textbf{49.63}} & \fst{\textbf{14.01}} & \fst{\textbf{0.370}} &  \fst{$\approx$ 8s} \\
     
    \bottomrule

    \end{tabular}}
    \vspace{-0.3cm}
    \caption{\textbf{Quantitative comparison for 3D generation}. We compare \method{} with baselines for text-to-3D generation utilizing text prompts from \textit{T3Bench}.}
    \vspace{-0.3cm}
    \label{tab:stage2_comp}
\end{table*}

\begin{figure*}[ht!]
    \centering
    \includegraphics[width=0.95\linewidth]{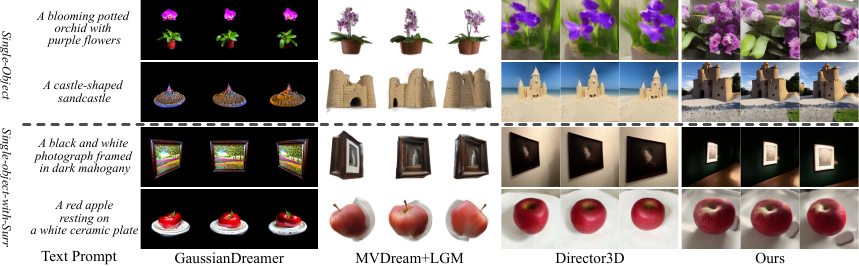}\vspace{-0.1cm}
    \caption{\textbf{Qualitative comparison for Stage2: Object-level 3D generation.} \method{} generates objects that align with the given description, incorporating rich background information and intricate details.}
    \label{fig:stage2_qualitative}
\vspace{-0.3cm}
\end{figure*}

\subsection{Implement Details}

During GS-VAE training (Stage 1), we set both the number of input and novel views to  \(N=4\)  for each multi-view scene. To improve model generalizability, we also sample 2 single-view images alongside the multi-view ones, applying the loss to the input views only for single-view images. The GS-VAE was trained on 8 A800 GPUs with a batch size of 32. The final model is trained for 200,000 iterations with approximately 4 days. We use gsplat~\cite{ye2023gsplat} as our 3D Gaussian renderer. 
We initialize the weights of our cross-view transformer from a pretrained RayDiff~\cite{zhang2024raydiffusion}.

For the MV-LDM (Stage 2), we employ Stable Diffusion 2.1~\cite{rombach2021highresolution} as our base model. During training, we set \(N=8\) for each multi-view scene. Similarly to Stage 1, we sample \(M=4\) single-view images alongside multi-view ones. For each iteration, we sample a batch size of 8 on each GPU. The final MV-LDM model was trained on 32 A800 GPUs, resulting in a total batch size of 3072 images. The model underwent 350,000 iterations, which took about 7 days.
We utilize a DepthAnything-V2-S~\cite{depth_anything_v2} model to estimate depth map on the fly. To achieve classifier-free guidance during sampling, we randomly drop text condition $\bt$ and pose condition $\bp$ with the probability 10\% during training.

\subsection{Evaluation Protocols}
\noindent\textbf{3D Reconstruction (Stage-1).} 
To demonstrate the 3D reconstruction generalization of our GS-VAE, we employed \textit{Tartanair}~\cite{wang2020tartanair} for our evaluation, which is a diverse synthetic dataset with 18 scenes not included in our training set,
covering both indoor and outdoor scenarios. 
Based on the degree of overlap and distance among input views, we categorize them into three distinct modes: \textit{Easy}, \textit{Medium}, and \textit{Hard}, each comprising 4 context views and 3 target views. 

We use the metrics PSNR, SSIM~\cite{wang2004image} and LPIPS~\cite{zhang2018unreasonable} for evaluating the reconstructed images. To better compare the reconstructed geometry, we evaluate the rendered depth maps. Follow~\cite{shao2024learning,hu2024depthcrafter}, we align rendered depth maps with the ground truth with per-scene scale and shift and calculate two widely recognized metrics~\cite{Ranftl2022midas} for evaluation, Absolute Mean Relative Error (\textit{AbsRel}) and $\delta$1 accuracy with a specified threshold value of 1.25.

\begin{figure*}[ht!]
    \centering
    \includegraphics[width=0.95\linewidth]{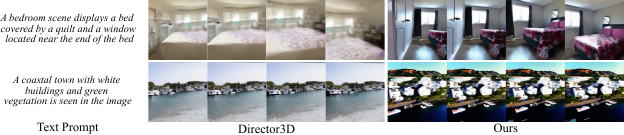}\vspace{-0.3cm}
    \caption{\textbf{Qualitative comparison for Stage 2: Scene-level 3D generation} with diverse scene-level prompt. Our result better aligns with the text prompt and captures more details.}
    \label{fig:stage2_qualitative_2}
\vspace{-0.5cm}
\end{figure*}

\noindent\textbf{3D Generation (Stage-2).} 
To assess the text-to-3D generation capabilities of our model, we employ two text prompt sets from \textit{T3Bench}~\cite{he2023t} –– \textit{Single-Object} and \textit{Single-Object-with-Surroundings}. These sets collectively evaluate the model's proficiency in object-level, and scene-level generation tasks. Additionally, we collected 80 diverse scene-level text prompts covering indoor and outdoor scenarios.

The quantitative results are evaluated using the CLIP-Score~\cite{hessel2021clipscore}, NIQE~\cite{mittal2012making}, and BRISQUE~\cite{mittal2012no} metrics. CLIP-Score assesses the alignment between the generated images and the textual prompts, whereas NIQE and BRISQUE indicate the image quality.

\subsection{Comparison with Baselines}
\noindent\textbf{3D Reconstruction.} 
We compare the GS-VAE of \method{} against two sparse-view reconstruction models –– namely, pixelSplat~\cite{charatan23pixelsplat} and MVSplat~\cite{chen2024mvsplat}. Our quantitative findings are shown in~\cref{tab:stage1_comp}, where we accentuate the \colorbox{firstcolor}{best}, \colorbox{secondcolor}{second-best}, and \colorbox{thirdcolor}{third-best} scores across all metrics. Beginning with geometry, \method{} surpasses the other two baselines, and this advantage becomes more pronounced as the degree of overlap among input views diminishes –– see the $\delta$1, which exhibits a relative enhancement of 44\% on \textit{Easy} mode and a substantial 64\% on \textit{Hard} mode against pixelSplat. Regarding the reconstructed images, \method{} delivers comparable outcomes on \textit{Easy} mode and notably outperforms its counterparts as the mode intensifies, particularly in \textit{Hard} mode. We also provide qualitative results in~\cref{fig:stage1_qualitative}. 
These findings on geometry and image reconstruction suggest that \method{} is more robust to variations in input view overlap than its baselines, a key factor for the success of our downstream 3D generation task.

\noindent\textbf{3D Generation.} 
We compare \method{} with three text-to-3D baseline methods, covering both optimization-based method and feed-forward method. GaussianDreamer~\cite{yi2023gaussiandreamer} is a state-of-the-art SDS-based method for 3DGS. We additionally implement a baseline that applies a multi-view to 3D method, LGM~\cite{tang2024lgm} to images generated by MVDream~\cite{shi2023MVDream}. We also compare to a feed-forward method, Director3D~\cite{li2024director3d} (without refiner). ~\cref{fig:stage2_qualitative} and \cref{fig:stage2_qualitative_2} show that our method is capable of generating both, object and scene-level contents, containing background and rich details, outperforms both optimization-based and feed-forward baselines. \cref{tab:stage2_comp} shows our overall metrics are suboptimal compared with Director3D for object level while leading in other cases. This is attributed to failure cases in the object-centric setting, see supplementary material for more details. Note that our method takes only 8 seconds for generation, outperforming all baselines.

\subsection{Ablations of GS-VAE (Stage 1)}
\label{subsec:stage1_ablation}

\begin{table}[t!]
    \centering
    \renewcommand{\tabcolsep}{6pt}
    \resizebox{0.98\linewidth}{!}{
    \begin{tabular}{@{}l|ccccc}
    
    \toprule
    \multirow{2}{*}{Variants} &  
    \multicolumn{5}{c}{Tartanair (4 views)} \\
    
    & 
    PSNR$\uparrow$ & SSIM$\uparrow$ & LPIPS$\downarrow$ & 
    \textit{AbsRel}$\downarrow$ & $\delta$1$\uparrow$  \\
    \midrule
    
    Ours(w/o RGB-D) &
    \trd{18.38} & \trd{0.475} & \snd{0.383} & \trd{0.761} & \trd{0.324} \\
    Ours(w/o single-view) &
    \snd{18.63} & \snd{0.480} & \trd{0.424} & \snd{0.542} & \snd{0.475} \\
    Ours &
    \fst{\textbf{19.49}} & \fst{\textbf{0.532}} & \fst{\textbf{0.341}} & \fst{\textbf{0.526}} & \fst{\textbf{0.505}} \\
     
    \bottomrule

    \end{tabular}}
    \vspace{-0.3cm}
    \caption{\textbf{Quantitative ablation results of GS-VAE} for generalizable reconstruction.} 
    \vspace{-0.4cm}
\label{tab:stage1_ablation}
\end{table}

\begin{figure}[t!]
  \centering
  \includegraphics[width=0.95\linewidth]{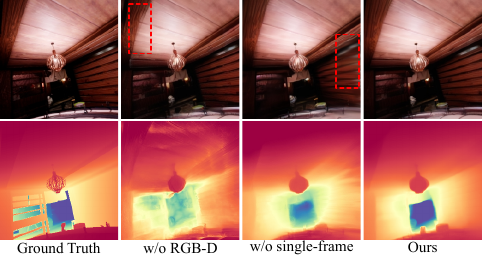}\vspace{-0.3cm}
  \caption{\textbf{Qualitative ablation results of GS-VAE} for generalizable reconstruction.}
    \label{fig:stage1_ablation}
\vspace{-0.5cm}
\end{figure}

In~\cref{tab:stage1_ablation}, we run ablation studies on the \textit{Hard} mode of \textit{Tartanair}, analyzing the following factors of GS-VAE.

\noindent\textbf{The effectiveness of Depth Prior for GS-VAE.}
We investigate the impact of RGB-D latent space during our stage 1 training. Our result in~\cref{tab:stage1_ablation} highlights that training without RGB-D latent space –– only RGB latent space –– yields worse results over our full model on geometry. Additionally, this bad geometry will lead to sub-optimal quality on reconstructed images, see qualitative results in~\cref{fig:stage1_ablation}.

\noindent\textbf{If large scale matters in Generalizable Reconstruction.} 
Next, we ablate the effectiveness of large-scale dataset, denoted as \textit{w/o single-view} in~\cref{tab:stage1_ablation}. The exclusion of the single-view dataset results in diminished performance across both reconstructed images and geometry. This underscores the significant role of large-scale datasets in achieving robust Generalizable Reconstruction. We also visualize the qualitative results in~\cref{fig:stage1_ablation}.

\subsection{Ablations of MV-LDM (Stage 2)}
\label{subsec:stage2_ablation}
\begin{table}[t!]
    \centering
    \renewcommand{\tabcolsep}{6pt}
    \resizebox{0.95\linewidth}{!}{
    \begin{tabular}{@{}ll|ccc@{}}
    \toprule
    & Variants & BRISQUE$\downarrow$ & NIQE$\downarrow$ & CLIP-Score$\uparrow$
     \\ 
    \midrule
    \multirow{2}{*}{\makecell{\textbf{Training} \\ \textbf{strategy}}}
     & w/o single-view data & \snd{59.45} & 14.57 & \trd{0.342} \\
     & w/o high noise level & 63.06 & \fst{\textbf{13.88}} & \snd{0.343} \\
     
    \midrule
    \multirow{2}{*}{\makecell{\textbf{Inference} \\ \textbf{strategy}}}
     & w/o hybrid sampling & \trd{66.19} & \snd{13.92} & 0.329 \\
     & w/o CFG-rescale & 89.70 & 15.15 & 0.303 \\
    \midrule
     & \underline{Ours (full)} & \fst{\textbf{58.88}} & \trd{14.00} & \fst{\textbf{0.369}} \\
    \bottomrule
    \end{tabular}}\vspace{-0.3cm}
    \caption{\textbf{Quantitative ablation results of MV-LDM} on text-to-3D generation.}
    \vspace{-0.6cm}
    \label{tab:stage2_ablation}
\end{table}

In~\cref{tab:stage2_ablation}, we conduct ablation studies on the \textit{Single-Object-with-Surroundings} subset of \textit{T3Bench},  examining both training and inference strategies of MV-LDM.

\noindent\textbf{Single-View Dataset.}  We assess the impact of single-view data. Excluding this training strategy by solely training with multi-view data (w/o single-view data) results in performance degradation, We attribute this to the lack of single-view data, which reduces the model's generalizability, aligning with the observations in MVDream~\cite{shi2023MVDream}.

\noindent\textbf{High Noise Level.} 
As mentioned in Zero123++~\cite{liu2023zero1to3} and \cref{sbusec:mvldm}, it is crucial for the model to learn high-level structures in the low-frequency space. Therefore, we also evaluate the effectiveness of a high noise level by setting $P_{mean} = -0.5$ and $P_{std} = 1.2$ during multi-view training (w/o high-noise level). The results in~\cref{tab:stage2_comp} show that both visual quality and CLIP score decline in this setting.

\noindent\textbf{Hybrid Sampling and CFG-Rescale.} Finally, we evaluate the design of our inference strategy~\cref{sbusec:mvldm} on hybrid CFG sampling and CFG-rescale by applying CFG solely on the text prompt(w/o hybrid sampling) and set the CFG-rescale factor to 0 (w/o CFG-rescale). \cref{tab:stage2_ablation} shows that the absence of hybrid sampling and CFG-rescale results in varying degrees of metric decline.

\section{Conclusion}
\label{sec:conclusion}

We present \method{}, a 3D-aware latent diffusion model tailored for text-to-3D generation at both object and scene levels in seconds. 
We demonstrate the effectiveness of our method in feed-forward reconstruction and 3D generation with extensive experiments.
We believe our work offers valuable contributions to text-to-3D scene generation, improving generalizability, fidelity, and efficiency.

{\small
\bibliographystyle{ieee_fullname}
\bibliography{bibliography_long,bibliography_custom}
}

\clearpage

\newpage
\onecolumn
\appendix
\clearpage
\setcounter{page}{1}

\twocolumn[{%
		\renewcommand\twocolumn[1][]{#1}%
		\maketitlesupplementary
            \vspace{-0.3cm}
		\begin{center}
			\includegraphics[width=0.999\textwidth]{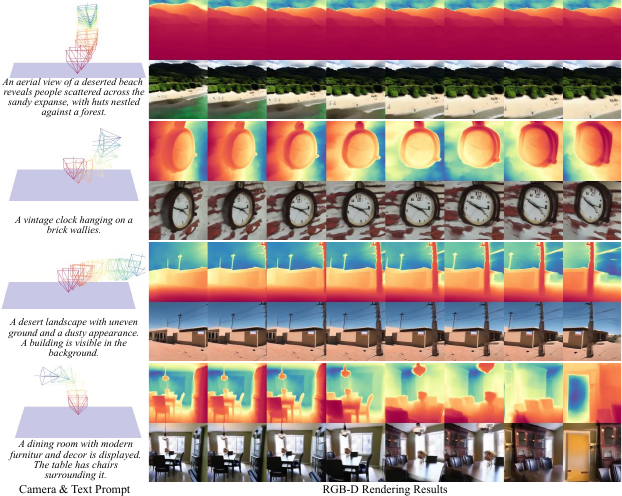}
   \captionsetup{type=figure}
			\captionof{figure}{
				\textbf{More Results.} Our method can synthesize diverse results across multiple domains, taking text prompts and camera poses as input. As shown in the image, we can render diverse (indoor/outdoor/object-centric) scenes that are faithfully aligned with the given text prompt and camera trajectory, while maintaining good underlying geometry.
			}
			\label{fig:moreresults}
		\end{center}
	}]

\section{More Generation Results}
\begin{figure*}[htpb]
  \centering
  \includegraphics[width=1\linewidth]{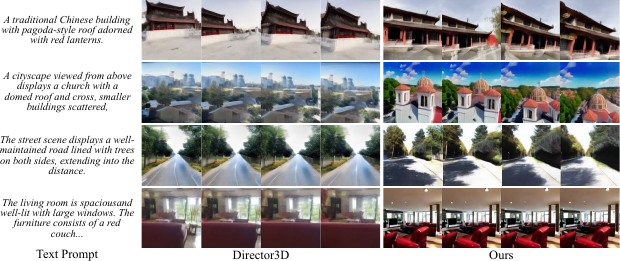}
  \vspace{-0.1cm}
  \caption{\textbf{Qualitative comparison with Director3D.} We compare \method{} against baselines under varying difficulty settings. As overlap gradually decreases, the advantages of our method continue to grow. Moreover, as shown in the depth map, our method exhibits superior geometry quality across all settings.}
    \label{fig:morewithdir3d}
\vspace{-0.3cm}
\end{figure*}

We present additional generation results of multi-view images and depth maps across diverse text prompts and camera trajectories in~\cref{fig:moreresults}. These results underscore the robustness of our approach in managing both object-level and scene-level prompts for 3D scene generation. Then we present more scene-level generation comparison results with Director3D, the concurrent scene-level feedforward text-to-3D method, as shown in~\cref{fig:morewithdir3d}.

\begin{figure*}[t!]
  \centering
    \includegraphics[width=1\linewidth]{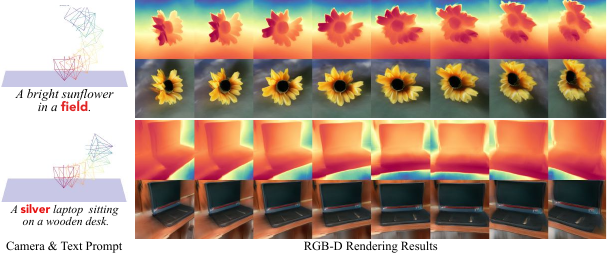}\vspace{-0.1cm}
  \caption{\textbf{Multiview-inconsistency cases.} We show Multiview-inconsistency, the main factor contributing to the failure cases of our method. As shown in the images, due to the lack of explicit 3D representation during multiview generation in latent space, \method{} will encounter view inconsistency under large rotations or extreme viewpoints.}
    \label{fig:limitation_view}
\vspace{-0.3cm}
\end{figure*}
\section{Limitations}

\begin{figure*}[t!]
  \centering
  \includegraphics[width=0.95\linewidth]{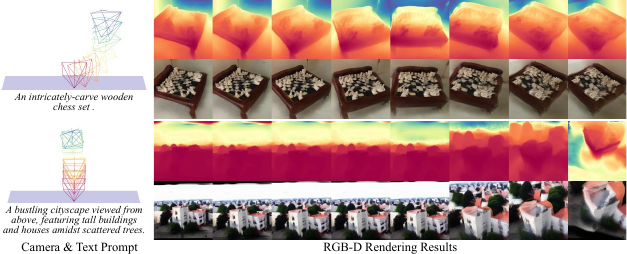}\vspace{-0.1cm}
  \caption{\textbf{Text-misalignment cases.} We then show Text-misalignment, the second factor contributing to the failure cases of our method. As shown in the images, \method{} synthesizes a black laptop instead of following the prompt, which should be silver.}
    \label{fig:limitation_text}
\vspace{-0.3cm}
\end{figure*}

We then visualize the failure cases of our method in~\cref{fig:limitation_text,fig:limitation_view}. Firstly, as shown in~\cref{fig:limitation_view}, despite specific designs during training and sampling aimed at mitigating 3D inconsistencies, \method{} still encounters inaccuracies in rendering high-frequency structures. Secondly, as shown in~\cref{fig:limitation_text}, our method occasionally exhibits text misalignment issues. The primary cause is the joint training of single-view and multi-view models, which disrupts the original text embedding layer of the pre-trained image generation model. Designing a specialized architecture to preserve the text alignment capability of the pre-trained image generation model will address this issue.

\end{document}